\documentclass[9pt,twocolumn,a4paper]{article}
\usepackage[a4paper,top=0.75cm,bottom=0.75cm,left=0.75cm,right=0.75cm,marginparwidth=1cm]{geometry}

\usepackage{cvpr}
\usepackage{times}
\usepackage{float}
\usepackage{epsfig}
\usepackage{graphicx}
\usepackage{amsmath}
\usepackage{amssymb}
\usepackage{url}
\usepackage{multirow}
\usepackage{arydshln}
\usepackage{hyperref}
\usepackage{caption}
\usepackage{subfigure}
\usepackage[nocompress]{cite}
\hypersetup{
    colorlinks=true,
    linkcolor=blue,
    filecolor=blue,      
    urlcolor=blue,
    citecolor=blue,
}
\usepackage[font=small,skip=3.5pt]{caption}

\cvprfinalcopy 

\def\cvprPaperID{****} 

\begin{document}

\title{\LARGE End-to-End Parkinson Disease Diagnosis \\
using Brain MR-Images  by 3D-CNN \\
\vspace{25 pt}}
\author{Soheil Esmaeilzadeh\\
{\small  Stanford University} \\ 
{\small soes@stanford.edu}
\and
Yao Yang\\
{\small  Stanford University} \\ 
{\small yangyao@stanford.edu}
\and
Ehsan Adeli\\
{\small  Stanford University} \\ 
{\small eadeli@stanford.edu}
}

\maketitle
%

\begin{abstract}
   In this work, we use a deep learning framework for simultaneous classification and regression of Parkinson disease diagnosis based on MR-Images and personal information (i.e. age, gender). We intend to facilitate and increase the confidence in Parkinson disease diagnosis through our deep learning framework.  
\end{abstract}

\section{Introduction}
Parkinson's disease (PD) is a long-term degenerative disorder of the central nervous system that affects the motor system \cite{pd1}. Symptoms of Parkinson's disease include shaking, rigidity, slowness of movement, difficulty with walking, thinking and behavioral problems, dementia, and etc. In 2015, PD affected 6.2 million people and resulted in about 117,400 deaths globally \cite{pd2}. The disease typically occurs in people who are over 60 years old, of which one percent are affected. The cause of Parkinson disease is not yet known, but mostly is believed to be due to genetic and environmental reasons \cite{pd3}. 
\section{Background and Related Works}
Computer aided diagnosis is getting common in healthcare recently \cite{Salv} \cite{soheil}. The diagnosis process for PD includes considering medical history and neurological examination. Computed tomography (CT) scans of people with PD usually appear normal and MRI has become more accurate in diagnosis of PD, especially through iron-sensitive T2 and SWI sequences at a magnetic field strength of at least 3T, both of which can demonstrate absence of the characteristic \textit{swallow tail} imaging pattern in the dorsolateral substantia nigra. There's a 98\% sensitivity and 95\% specificity for PD on the absence of the pattern.
Brain MR-Images (MRI) are theeqrefore widely used for diagnosing Parkinson in order to improve patient treatment strategies. MR-Images are also used to rule out other diseases that can be secondary causes of parkinsonism, most commonly encephalitis, and chronic ischemic insults. Diagnosing diseases based on radiologists' reading on MRI images are oftentimes prone to mistakes. In recent years, machine learning methods become a common tool for  early-stage diagnosis to localize the disease in the brain (i.e., localization of disease markers). Salvatorec et al. used a machine learning algorithm that allows individual differential diagnosis of PD by means of MRI which is able to obtain voxel-based morphological biomarkers of PD \cite{Salv}. Zhang et al. used a machine learning framework based on principal components analysis (PCA) and Support Vector Machine (SVM) for the classification of Parkinson's disease and Essential Tremor (ET). They used statistical analysis and machine learning method to test the differences between PD and ET in some specific brain regions \cite{zhang}. Another school of research focuses on Region of Interest methods (ROI) where some specific regions of the brain such as the gray matter, hippocampal volume, and cortical thickness are extracted due to a priori knowledge about their effects on brain functionality
and memory \cite{Liu} \cite{Peng}. 
\subsection{Baseline}
For diagnosis of Parkinson disease using MR-Images the state-of-the-art works done by Ahmed et al. \cite{Ahmed} and Gil et al.\cite{Gil} serve as the baselines for our model accuracy. The best performance reported by Ahmed et al. \cite{Ahmed} using an ANN model is 70 percent, and by Gil et al. \cite{Gil} combining the ANN and SVM classifier has an accuracy level of 86.96 percent where both of them use human-engineered feature extraction for training the models. 
\subsection{Motivation}
As mentioned before, most of the current methods focus on using human-engineered features extracted from MRI data, which have less optimal learning performances because of the possible high correlation between engineered features and the subsequent classification or regression models. We believe that integrating the feature extraction and the learning of models into one framework can improve the diagnostic performance. Moreover, by extracting the brain's heat-map after the training process, we will be able to find the important parts of the brain that contribute to the Parkinson disease, and this finding can serve as a valuable information for Parkinson diagnosis by medical practitioners. Hence, with an end-to-end approach and without using apriori human-engineered feature extraction techniques we use MR-Images data in order to classify them as Parkinson Disease (PD) or Healthy Condition (HC) and we use the best accuracy found in the previous works, 86.96 percent, as the baseline for Parkinson disease classification. As the input, we use three-dimensional brain MR-Images (section \eqref{sec_Data}), and use a Convolutional Neural Network as the training model (section \eqref{sec_Model}).

\section{Dataset and Features} \label{sec_Data}
\begin{table}[t!]
\begin{center}
\resizebox{0.49\textwidth}{!} {
\begin{tabular}{c | c c c c   c c c c c}
\hline
\multirow{2}{*}{Sex} & \multirow{2}{*}{Group}  & \multicolumn{8}{c}{Age}                 \\ 
&    & count & mean & std & min & 25\%  & 50 \%  & 75 \%  & max  \\ 
\hline \hline
F  & HC & 70 & 59.2  & 11.6 & 31 & 53 & 60 & 68 & 84   \\                       
F  & PD & 160 & 61.9 & 9.9  & 35& 56 & 62 & 69& 84     \\
\hdashline
M   & HC & 134 & 61.7 & 10.9 & 31 & 57 & 63& 69 & 83   \\
M   & PD & 292 & 63.3 & 9.8 & 36 & 57 & 64& 71 & 89   \\
\hline
\end{tabular}
}
\end{center}
\caption{Statistical overview of the MR-Images data; HC: Healthy condition, PD: Parkinson disease}
\label{tab:1}
\end{table}

\subsection{Format of Data}
In this work, a set of three-dimensional brain MR-Images is used for Parkinson Disease diagnosis . The dataset for this work is from the PPMI database \cite{Marek}. Fig. \eqref{f_MRI1} shows an illustration of an MR-Image cut in three sagittal, coronal, and axial planes respectively with cut coordinates of $x=36, y=10, z =36$ where brain tissues together with skull, scalp, and dura are observed. The size of MR-Image is $(120,120,270,1)$ as MR-Image is gray-scale leading to a size of 4 million pixels for each image that will serve as visual features.



\subsection{Statistics of Data}
\begin{figure*}
  \centering
  \subfigure[Age-Frequency distribution]
  {\includegraphics[scale=.36]{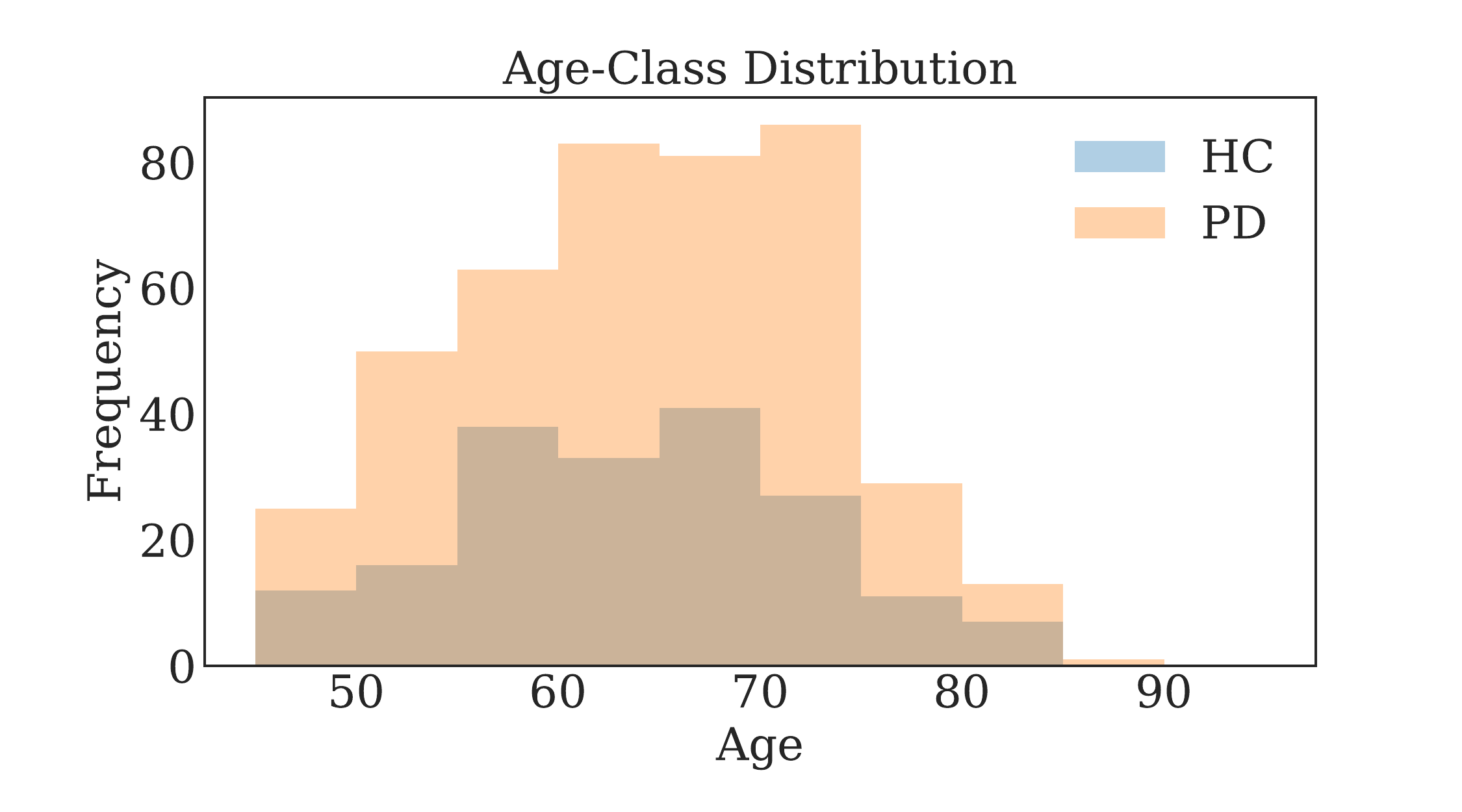}  \label{f1_1} }\quad
  \subfigure[Age-Class distribution]
  {\includegraphics[scale=.45]{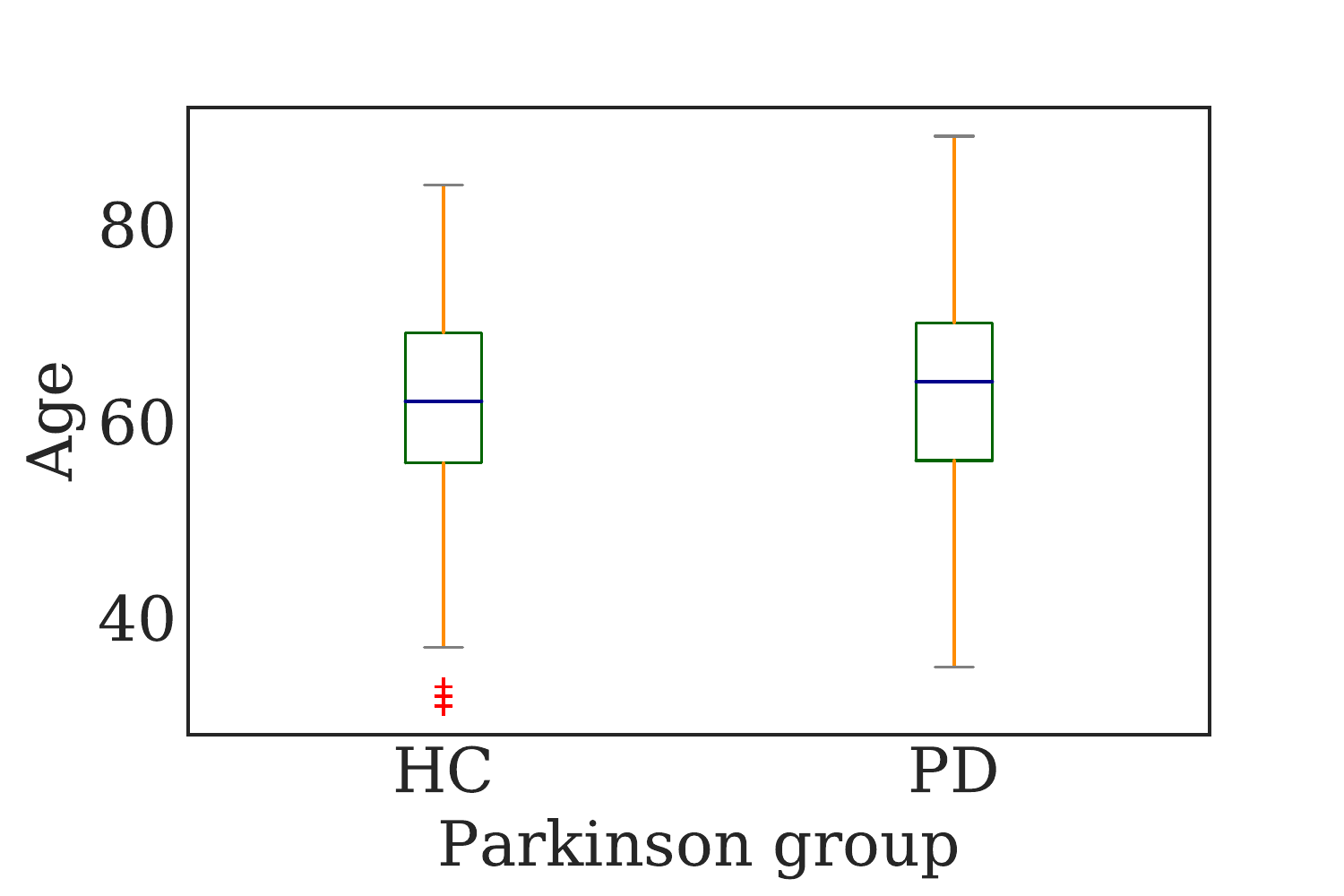}  \label{f1_2} }
 \caption{Parkinson disease dataset overview; HC: Healthy condition, PD: Parkinson disease}
\end{figure*}

We report the demographic information of subjects in Table \eqref{tab:1}. The dataset consists of 452 Parkinson patients (PD), including 292 males (M) and 160 females (F) and 204 images from people in healthy conditions (HC) with 134 males and 70 females. The average age of the patients is 61 where the minimum age is around 30 and the maximum age is 89 (cref. Table \eqref{tab:1}). Fig. \eqref{f1_1} shows the number of patients in different classes of Parkinson disease as a function of age and Fig. \eqref{f1_2} shows the age distribution of different classes of Parkinson disease. The average age of the patients is around 61 in each class of disease (cref. Fig. \eqref{f1_2}) among male/female and we include age and gender as extra features besides MR-Images in training of our model. Besides, both male and female groups have approximately similar portions of patients in each of the Parkinson classes; in spite of this, we consider gender also as a feature in training the model in addition to age. Hence, the MR-Images, age, and gender of patients are the features that are used for training of our deep learning model that is presented in section \eqref{sec_Model}. \\

\subsection{Preprocessing of Data}
\subsubsection{Skull-Stripping}
\begin{figure*}[!htb]
  \centering
  \subfigure[ ]
  {\includegraphics[scale=.45]{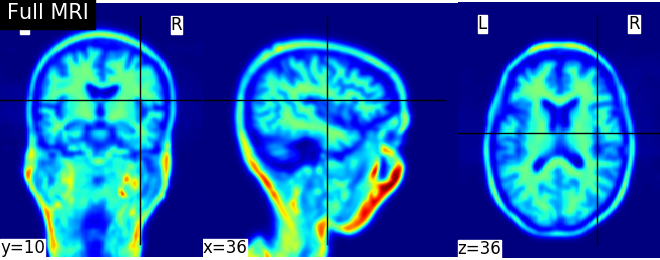}  \label{f_MRI1} }\quad
  \subfigure[ ]
  {\includegraphics[scale=.45]{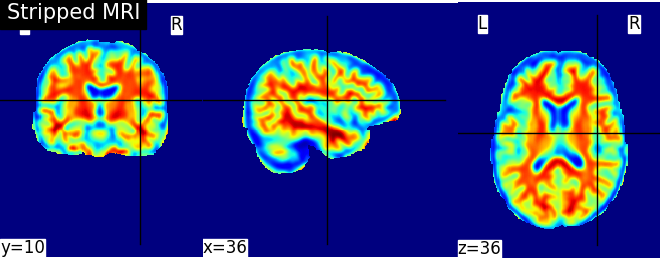}  \label{f_MRI2} }
 \caption{(a) full and (b) skull-stripped brain MR-Image - from \textit{left} to \textit{right}: Coronal, Axial, and Sagittal views}
\end{figure*}
In the preprocessing stage, we carry out skull-stripping to remove non-cerebral tissues like skull, scalp, and dura. In Fig. \eqref{f_MRI2} a skull-stripped version of MR-Image is given. Briefly, skull stripping acts as the preliminary step in numerous medical applications as it increases the speed and accuracy of diagnosis and includes removal of non-cerebral tissues like skull, scalp, and dura from brain images. Skull stripping can be part of the tissue segmentation (e.g. in SPM) but is mostly done by specialized algorithms that delineate the brain boundary. See \cite{Boesen2004} for a comparison of some brain extraction algorithms (BSE, BET, SPM, and McStrip), which suggests that all algorithms perform well in general but results highly depend on the particular dataset. \\
In our work we use the Brain Extraction Technique (BET) proposed by Smith in 2002 \cite{Smith2002} together with a Statistical Parametric Mapping (SPM) as a voxel-based approach for brain image segmentation and extraction and choose the stripped version with higher brain tissue intensity. Performing skull-stripping on brain MR-Images reduces the size of MR-Images to $(80,100,108,1)$ (i.e. 800,000 pixels), leading to a reduction factor of 4.61. 
\subsubsection{Data Augmentation}
Furthermore, we perform a data-augmentation technique on the MR-Images to expand the training set size. For this purpose, in the skull-stripped MR-Images we flip the right and left hemispheres of a brain for each patient and keep everything else the same. By doing this, we double the size of our dataset and can further train our model on a larger dataset than the original available samples.

\section{Machine Learning Model} \label{sec_Model}
\begin{figure*}
\centering
  \includegraphics[width=40pc]{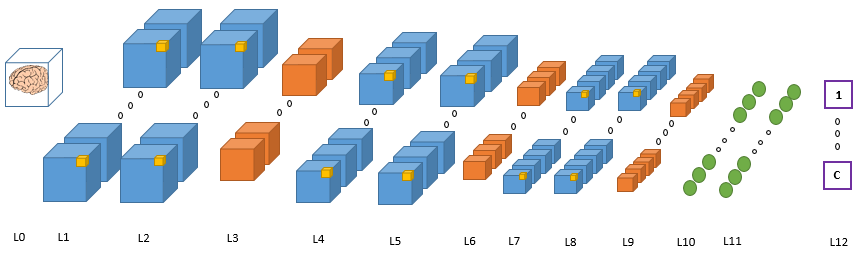}
  \caption{Architecture of the 3D-Convolutional Neural Network model: L$_0$: MR-Image (80$\times$100$\times$108$\times$1); L$_{1,2}$: Conv. ($3^3 \times 32)$; L$_{4,5}$: Conv. $(3^3 \times 64)$; L$_{7,8}$: Conv. ($3^3 \times 128$); L$_3$, L$_6$, L $_9 $: Max-pool ($2^3$ ,$4^3$ ,$4^3$); L$_{10}$, L$_{11}$: F.C.(512 ,128); L$_{12}$: Output (c) - \textit{same} padding for L$_{1,2,4,5,7,8}$, and strides of two for max-pooling layers}
  \label{fig_3dcnn}
\end{figure*}

In this work, we divide the dataset into a training ($85 \%)$, a development ($10 \%)$ and a test set ($5\%)$, and split them into batches of size 8 (due to memory issues we cannot test the effect of different batch sizes on the training process and just use size of 8). During the training, process we keep track of training and development set accuracies and loss function values. F$_2$-score (Eqs. \eqref{eq_prec}-\eqref{eq_Fbeta}) is being used which weighs recall higher than precision (by placing more emphasis on false negatives) to evaluate the performance of our model with true positive, true negative, false positive, and false negative being as TP, TN, FP, and FN respectively.
\begin{equation}
  \text{Precision} =\dfrac{TP}{TP+FP} \\ 
  \label{eq_prec}
\end{equation}
\begin{equation}
  \text{Recall} =\dfrac{TP}{TP+FN}
  \label{eq_recall}
\end{equation}
\begin{equation}
  F_2 = \dfrac{5~\text{precision}  \times \text{recall}}{4~\text{precision}+\text{recall}}
  \label{eq_Fbeta}
\end{equation}
%
For training process, we build a 3D Convolutional Neural Network (3D-CNN) shown in Fig. \eqref{fig_3dcnn} and integrate the feature extraction and the learning of the model into a unified framework. 
In the 3D-CNN shown in Fig. \eqref{fig_3dcnn}, $L_0$ is the input layer, which is the 3D MR-Image of brains with input dimension of $(80,100,108)$. $L_{1,2,4,5,7,8}$ layers are the conv-layers with \textit{stride} of 1 and \textit{same} padding and $L_{3,6,9}$ are the max pooling layers with \textit{stride} of 2 (dimensions are given in Fig. \eqref{fig_3dcnn}; strides and filter paddings are not tuned as hyperparameters in this work due to lack of enough computational resources). The proposed model repeats the blocks of 2-\textit{conv.} and 1-\textit{max-pooling} layers for three times which are followed with two fully connected layers and the final output layer. For each layer, we use Leaky-Rectified-Linear-Unit (Leaky-ReLU) as the activation function. For the output layer for Parkinson classification we use the Softmax classifier that uses the cross-entropy loss. For the optimization process we use Adam method \cite{adam} and all the implementations are done in \textit{TensorFlow} framework. \\ 
\\
For the training process of the model shown in Fig. \eqref{fig_3dcnn}, we name the model as the \textit{original model}, where two convolutional layers are followed by a max-pooling layer, repeated three times, and finally followed by two fully connected layers. We have also considered a sub-model of the one shown in Fig. \eqref{fig_3dcnn} where only one convolutional layer precedes each max-pooling layer, and a three conv-max-pool pairs of them are followed by two fully connected layers, and refer to this sub-model as the \textit{simplified model}, which its number of parameters are considerably lower than the \textit{original model} and lead to shorter training times. \\
\\
For each of the \textit{original model} and \textit{simplified model}, we perform a random search \cite{James} for hyper-parameter optimization. We look for optimum values of drop-out rate in the fully connected layers, regularization coefficients for kernel and bias in the 3D convolutional layers, $\alpha$ coefficient in the Leaky-ReLU activation function (Eq. \eqref{eq_leaky}), and learning-rate ($Lr$) of the optimizer. 
\begin{equation}
\text{Leaky-ReLU:~~}  f(x) =
\left\{
	\begin{array}{ll}
		x  & \mbox{if } x \geq 0 \\
		\alpha x & \mbox{if } x < 0
	\end{array}
\right.
\label{eq_leaky}
\end{equation}
Furthermore, we experiment an exponential learning-rate decay for the optimizer which has the mathematical form of $Lr = {Lr}_0~e^{-kt}$ where the initial learning rate (${Lr}_0$) and the decay steps ($t$) are hyper-parameters that are tuned in case of using decaying learning rate. \\
Following each convolutional layer, we put a normalization layer as well. For the normalization layer we have experimented batch normalization and group normalization \cite{gnorm} as two different approaches. Moreover, we investigated the effect of adding/removing gender and age of each patient to the model as two extra features in the last fully connected layer.

\section{Experiments and Results}
\begin{figure*}
  \centering
  \subfigure[ ]
  {\includegraphics[scale=0.4]{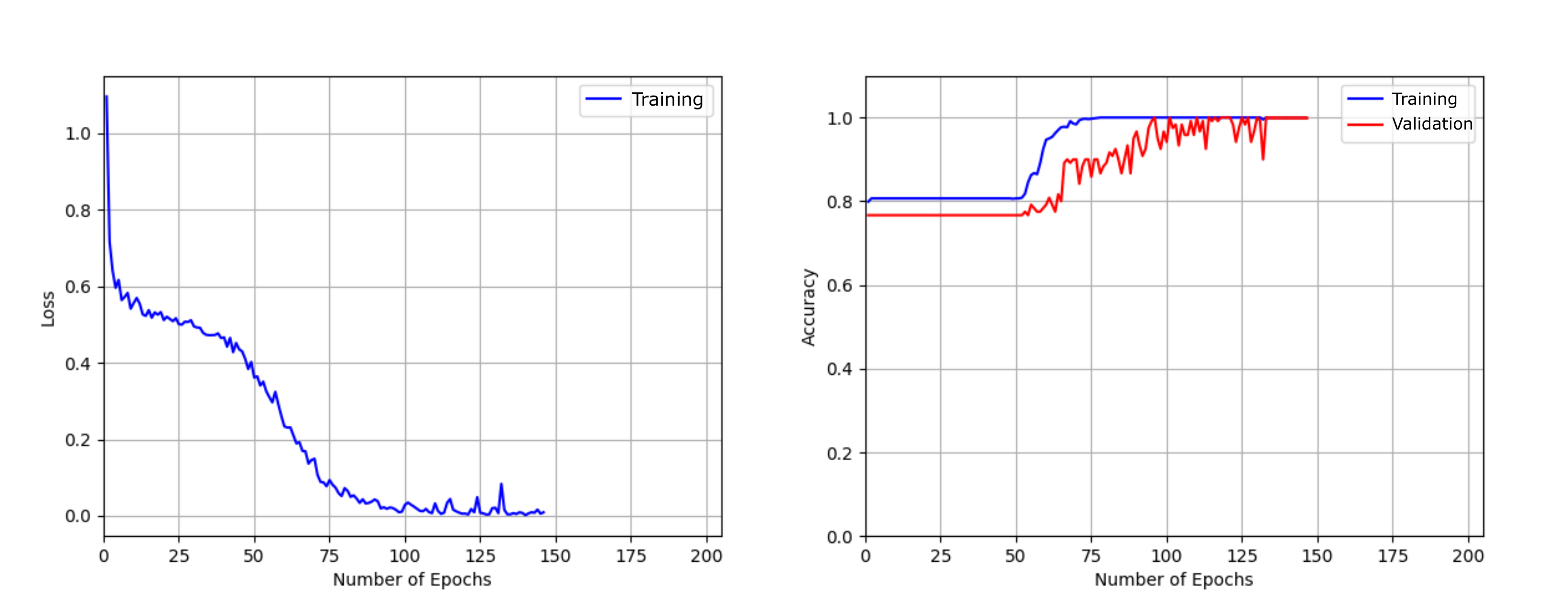}  \label{f_train_proc1} } \quad
 \caption{Loss value and accuracy of the training-set and validation-set during the training process}
\end{figure*}
\subsection{Experiments} \label{sec_experiments}
Tables \eqref{tab_exp1} and \eqref{tab_exp2} show the experiments and hypter-parameters study results. Starting with the \textit{original model} shown in Fig. \eqref{fig_3dcnn} we performed a hyper-parameter study to find the optimum value of learning rate (0.00005), and when the training accuracy reached 100 percent we got accuracy of 77.1 percent on the validation set. Repeating the similar procedure with the \textit{simplified model} described at the end of section \eqref{sec_Model} we reached to a higher validation accuracy i.e. 82 percent. This implies that probably the original model due to having too many parameters built in it might overfit the training data, leading to the lack of generalization and theeqrefor not performing well on the validation set. \\
In the next step of experiments, we added age and gender of the patients as two additional features to the last fully connected layer. By adding age and gender the validation accuracies of the \textit{original model} and \textit{simplified model} increased by 5\% and 2.5 \% respectively, leading to 81.2 and 84.1 percent accuracy. It was also observed the age and gender addition led to faster training process. Worth mentioning that, neglecting MR-Images and doing a simple logistic regression analysis on status of Parkinson with respect to patinets age led to 72\% accuracy. \\
Afterwards, we added normalization layers after the convolutional layers in both \textit{original} and \textit{simplified} models. As the normalization layers, we experimented the training with Batch Normalization and Group Normalization. In general, both type of normalizations led to slight increase in the validation accuracies and decrease in training time (100 \% train accuracy was achieved in a fewer number of epochs).\\
In the next set of experiments, we added bias and kernel regularizations to the layers of both \textit{simplified} and \textit{original} models and by hyper-parameter study we found the optimum regularization coefficients for each, shown in Table \eqref{tab_exp2}. Adding regularization, in both of models delayed the overfit on train dataset and led to further learning and higher validation accuracy in each. \\
Later on, we added drop-out, which as a regularization term affects the last two fully connected layers in both of the \textit{simplified} and \textit{original} models. Letting the keep probability rate ($1-$dropout rate) to vary for both of the fully connected layers, we carried out a parameter study to find the proper keep probability rates, which are not too big to lead to early overfit and not too small to inhibit learning. With the optimum values of regularization coefficient and keep probability rates shown in Table \eqref{tab_exp2} the simplified model which includes age and gender of the patients led to 100 \% accuracy on both the train and validation sets. We eqrefer to this as the best trained model. Just for the sake of illustration of the difference between using group and batch normalization, for the best trained model, the training loss and train and validation set accuracies as the function of number of epochs are shown in Fig. \eqref{f_train_proc1}. It can be seen that after training set accuracy reaches to 100 \% the model reaches to full accuracy on the validation set as well (100 \% validation accuracy), and ultimately the loss function reaches to zero values. This model, on the test-set of 56 samples also led to 100 \% test accuracy (see Fig. \eqref{f_conf3}) \\
For the best trained model found above, the classification accuracies on both groups of Healthy patients (HC) and Parkinson diagnosed patients (PD) is presented in section \eqref{sec_metrics} by confusion matrices and ROC curves. Moreover, a heat-map for sensitivity analysis of the best trained model's output using image occlusion technique is presented in section \eqref{sec_heatmap}.

\begin{table*}
\begin{center}
\resizebox{0.5\textwidth}{!} {
\begin{tabular}{c|c|c|c}
\hline
\multirow{2}{*}{No.} & \multirow{2}{*}{Experiment}          & \multicolumn{2}{c}{Accuracy}   \\ 
 &         								 	& Train.  & Val. 	\\ \hline  \hline  
 1& Original Model (OM)                  	& 1.0     & 0.771   	\\ \hline
 2& Simplified Model (SM)                  	& 1.0     & 0.820   	\\ \hline
 3& OM + Gender \& Age (OM-GA)             	& 1.0     & 0.812   	\\ \hline
 4& SM + Gender \& Age (SM-GA)              & 1.0     & 0.841   \\ \hline
 5& OM-GA + Batch Normalization (OM-GA-B)   & 1.0     & 0.821   \\ \hline
 6& SM-GA + Batch Normalization (SM-GA-B)   & 1.0     & 0.847   \\ \hline
 7& OM-GA + Group Normalization (OM-GA-G)  	& 1.0     & 0.820   \\ \hline
 8& SM-GA + Group Normalization (SM-GA-G)  	& 1.0     & 0.849   \\ \hline
 9& OM-GA-G~+~Regularization (OM-GA-GR)  	& 1.0     & 0.895   \\ \hline
10& SM-GA-G~+~Regularization (SM-GA-GR) 	& 1.0     & 0.935   \\ \hline
11& OM-GA-GR~+~Drop-out~(OM-GA-GRD)    		& 1.0     & 0.947   \\ \hline
12& SM-GA-GR~+~Drop-out~(SM-GA-GRD)    		& 1.0     & 1.000   \\ \hline
\end{tabular}
}
\end{center}
\caption{F$_2$-score values of different experiments on training set and validation set; OM: Original Model, SM: Simplified Model, GA: Gender \& Age included, B: Batch Normalization, G: Group Normalization, R: Bias \& Kernel Regularization, D: with Drop-out}
\label{tab_exp1}
\end{table*}
\begin{table}
\begin{center}
\resizebox{0.49\textwidth}{!} {
\begin{tabular}{c|c|c|c|c|c|c}
\hline

No.&  Experiment& 	$Lr$  & $\alpha$ & $Rc$ & $kp_1$ & $kp_2$ \\ \hline  \hline  
 1&   OM&  			0.00005      & 0 	&0 		& 1  	&1 	\\ \hline
 2&   SM&   		0.00005     & 0 	&0 		& 1  	&1 	\\ \hline
 3&   OM-GA&   		0.00020     & 0 	&0 		& 1  	&1 	\\ \hline
 4&   SM-GA&   		0.00005     & 0 	&0 		& 1  	&1 	\\ \hline
 5&   OM-GA-B&   	0.00005     & 0.01 &0 		& 1  	&1 	\\ \hline
 6&   SM-GA-B&   	0.00001     & 0.01 &0 		& 1  	&1 	\\ \hline
 7&   OM-GA-G&   	0.00001     & 0.01 &0 		& 1  	&1 	\\ \hline
 8&   SM-GA-G&   	0.00001     & 0.01 &0 		& 1  	&1 	\\ \hline
 9&   OM-GA-GR&   	0.00001     & 0.01 &0.05 	& 1  	&1 	\\ \hline
10&   SM-GA-GR&   	0.00001     & 0.01 &0.001 	& 1  	&1 	\\ \hline
11&   OM-GA-GRD&   	0.00001     & 0.01 &0.05 	& 0.2  	&0.35 	\\ \hline
12&   SM-GA-GRD&   	0.00001     & 0.01 &0.001 	& 0.45  &0.5 	\\ \hline
\end{tabular}
}
\end{center}
\caption{Parameters of different experiments of Table \eqref{tab_exp1}; $Lr$: Learning-rate, $\alpha$ Leaky-ReLU function's coefficient, $Rc$: Regularization coefficient, $kp_i$: keep probability of the $i-th$ fully connected layer (keep probability = $1-$dropout probability)}
\label{tab_exp2}
\end{table}
%
%
\subsection{Model Evaluation} \label{sec_metrics}
\begin{figure}
  \centering
  \subfigure[ ]
  {\includegraphics[scale=.3]{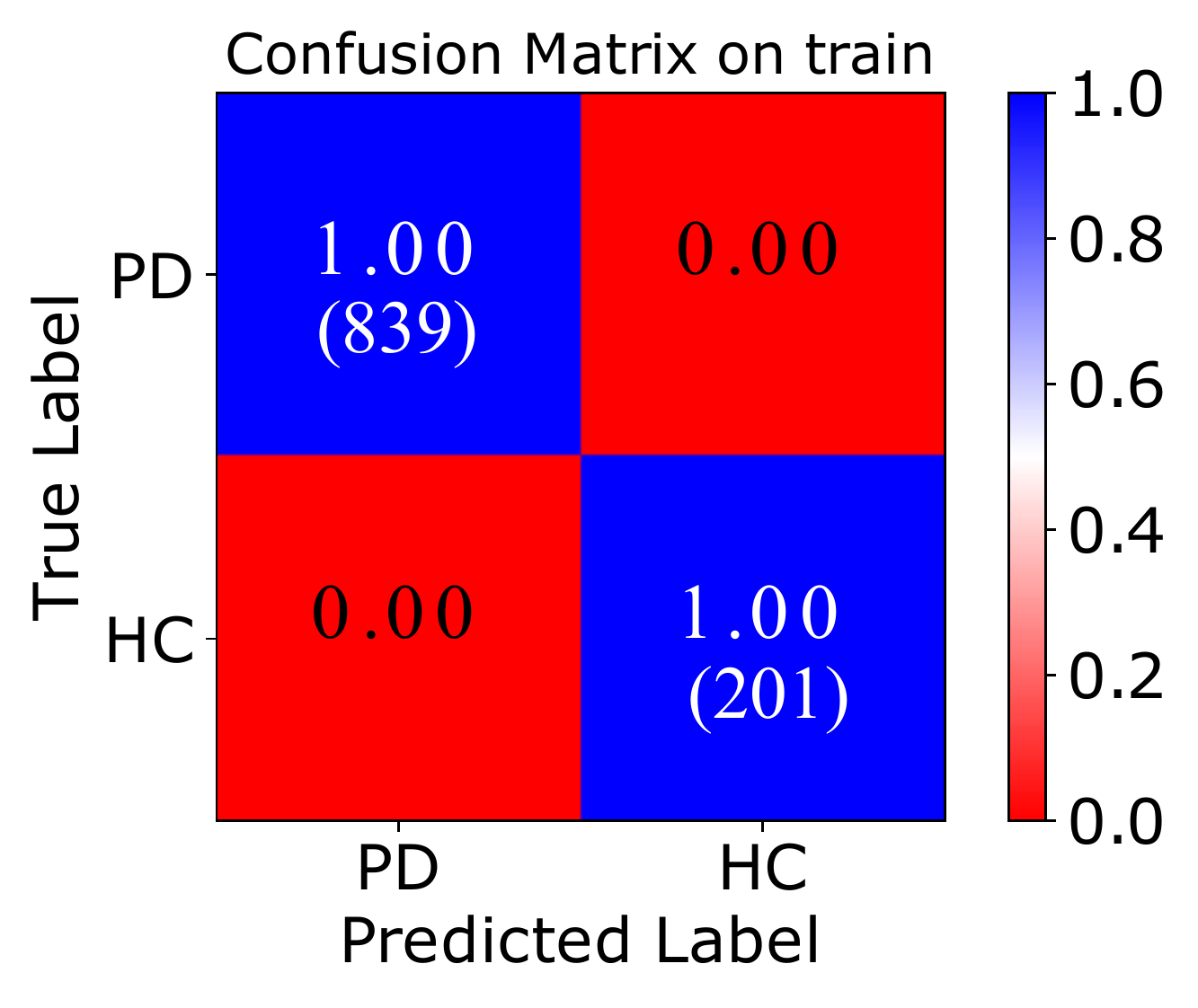}  \label{f_conf1} }\quad
  \subfigure[ ]
  {\includegraphics[scale=.3]{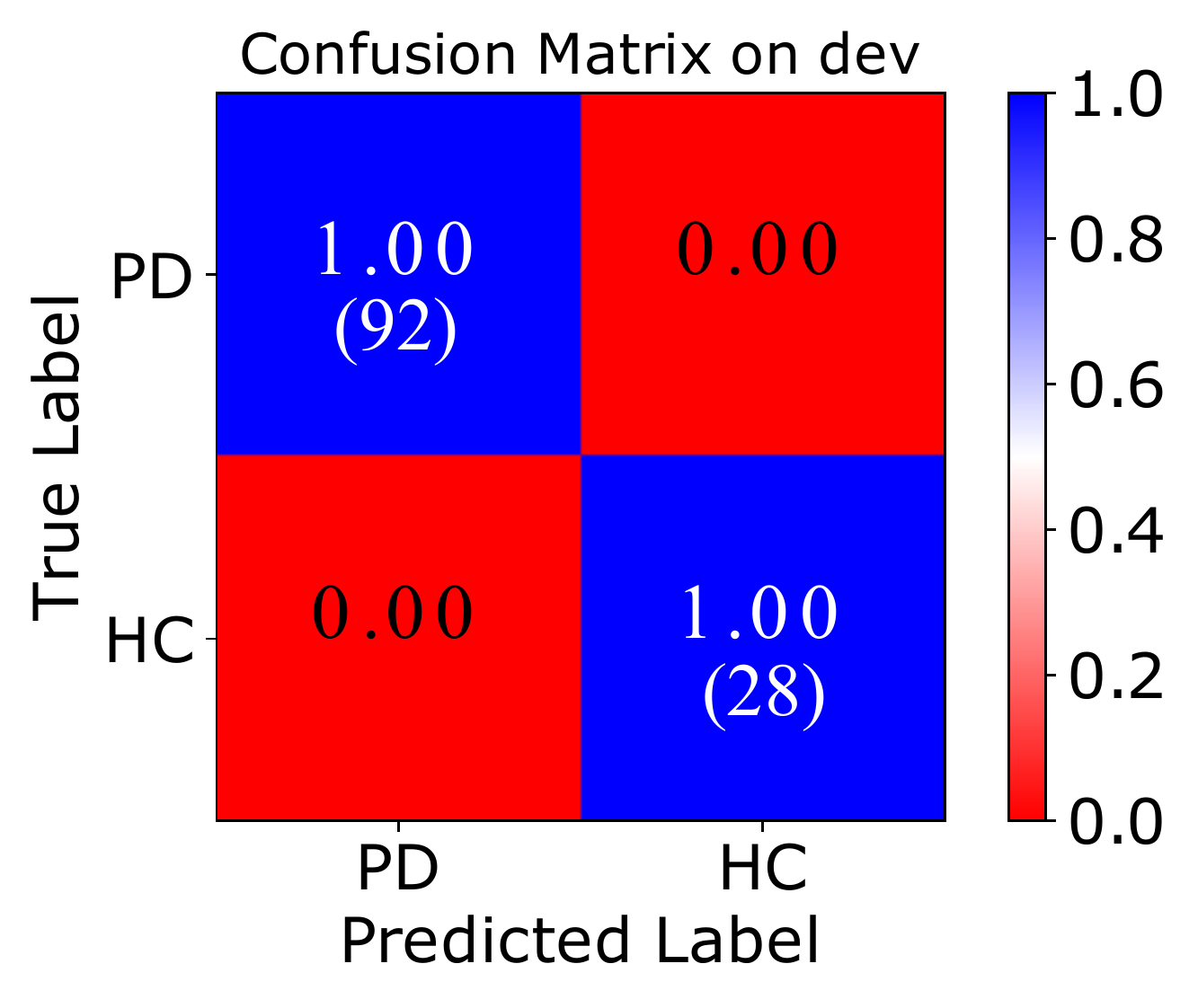}  \label{f_conf2} }\quad
  \subfigure[ ]
  {\includegraphics[scale=.3]{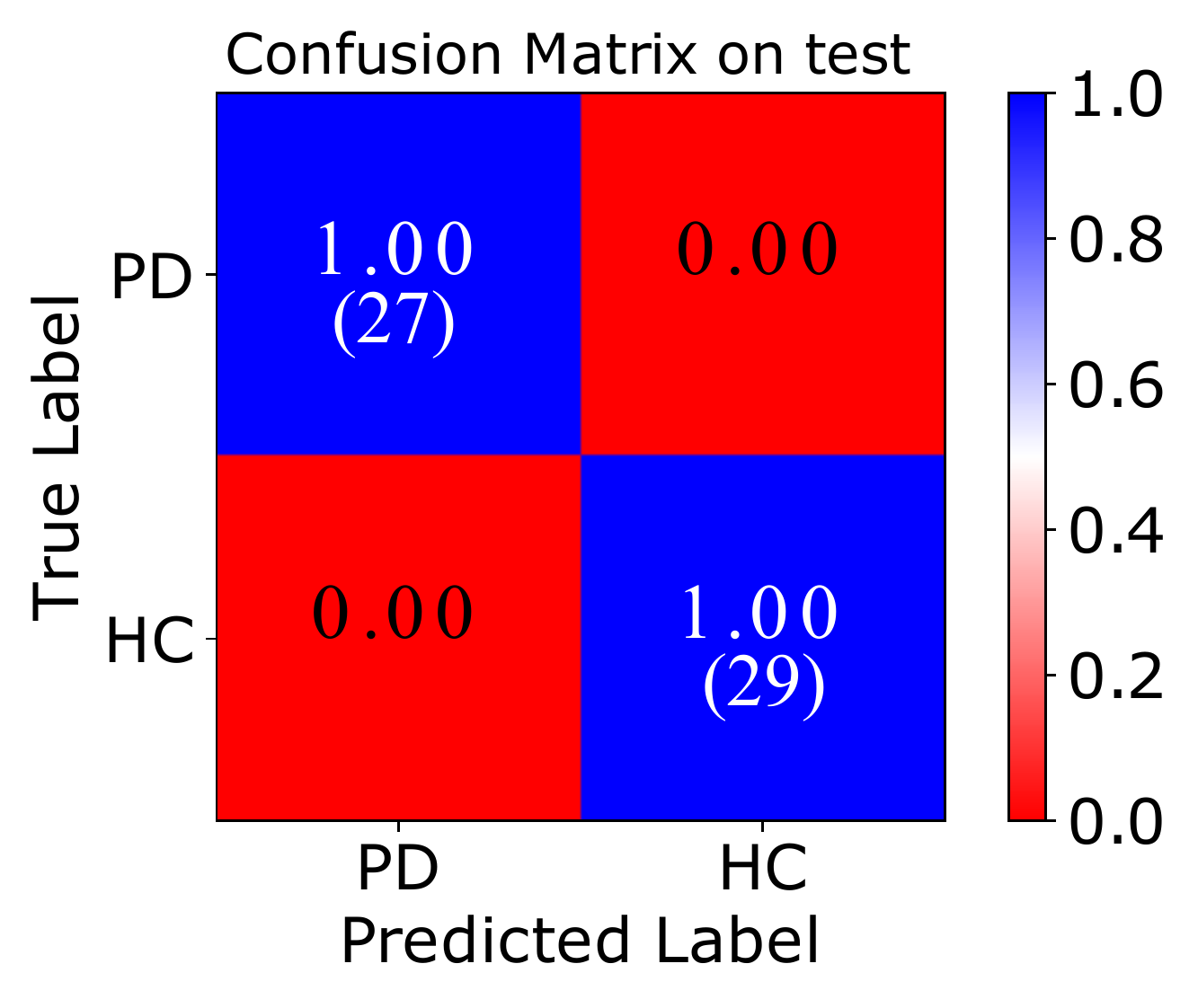}  \label{f_conf3} } 
 \caption{Normalized confusion matrix - \textit{top} to \textit{bottom}: train-set (1040 cases), dev-set (120 cases), test-set (56 cases) - HC: Healthy condition, PD: Parkinson disease}
\end{figure}

\begin{figure}
  \centering
  \subfigure[ ]
  {\includegraphics[scale=.49]{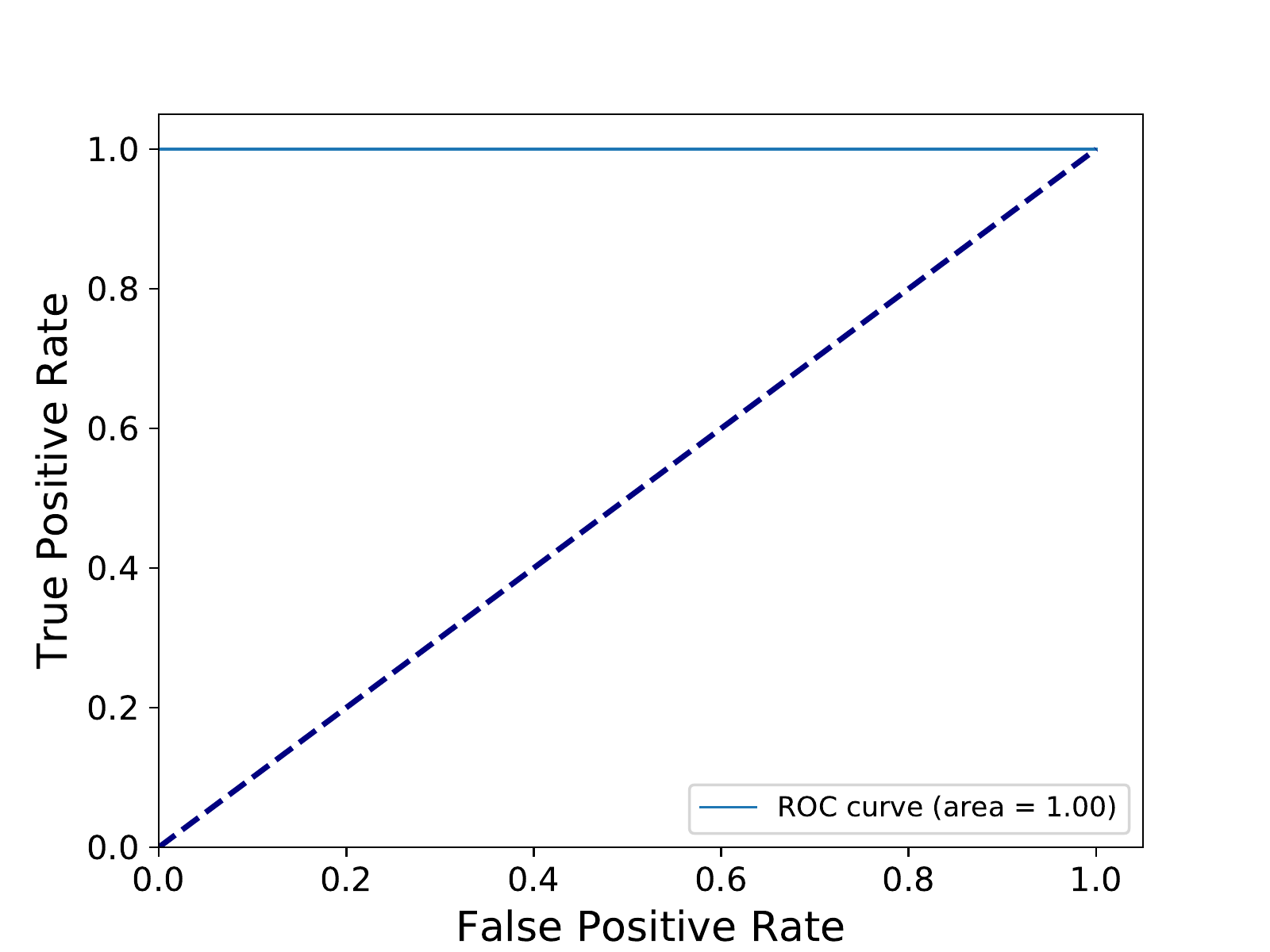}  \label{f_roc1} }\quad
 \caption{ROC curve for training-set (1040 cases), dev-set (120 cases), and test-set (56 cases)}
\end{figure}

In this part, to evaluate the performance of the best trained model found in section \eqref{sec_experiments}, we find the confusion matrix where each row of the matrix represents the instances in a predicted class while each column represents the instances in an actual class or vice versa. Figs. \eqref{f_conf1}, \eqref{f_conf2}, and \eqref{f_conf3} show the normalized confusion matrices of the Parkinson classification results of the best trained model for training-set, validation-set, and test-set respectively. As it can be seen the trained model performs pretty well on the validation and test sets (120 \& 56 cases respectively) with classifying the MR-Images as for Healthy Conditions (HC) and Parkinson Disease (PD) with accuracy of 100 percent, and leading to zero false negatives and zero false positives. \\ 
Fig. \eqref{f_roc1} shows ROC curves for training, validation, and test sets where the AUC is 1, shows perfect TPR v.s FPR.
\subsection{Parkinson Heat-Map of Brain} \label{sec_heatmap}
\begin{figure*}
\centering
  \includegraphics[width=0.75\textwidth]{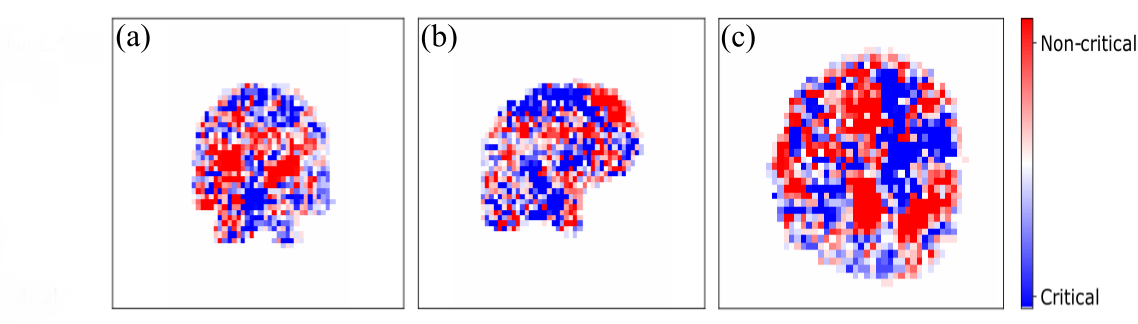}
  \caption{Brain's Heat-map for Parkinson diagnosis - from \textit{left} to \textit{right}: (a) Coronal, (b) Axial, and (c) Sagittal views}
  \label{fig_heatmap}
\end{figure*}

Similar to the approach in the work by Matthew et al. in 2013 \cite{Matthew2013} on visualizing and understanding Convolutional Networks, we perform a sensitivity analysis of the classifier output by occluding portions of the input image, revealing which parts of the brain are important for Parkinson diagnosis. For this reason, we performed an image occlusion analysis on our best-trained model found in section \eqref{sec_experiments} by translating a box with the size of $2\times 2\times 2$ zero-valued voxels along the whole MR-Image of a Parkinson patient that was correctly labeled as  Parkinson Disease (PD) by the trained model. In the heat-maps in Fig. \eqref{fig_heatmap} White areas are irrelevant as they don't change the confidence of the prediction, red areas increase the confidence, and blue areas decrease the confidence of the model suggesting that they are areas that are important for diagnosing of PD. \\
By looking at Fig. \eqref{fig_heatmap} in both Coronal and Axial views we see that \textit{Basal Ganglia} and \textit{Substantia Nigra} (bottom blue regions) together with \textit{Superior Parietal} part on right hemisphere of the brain are found to be of critical importance in diagnosis of Parkinson, where the former one is completely corroborated by medical studies that when dopamine receptors in the striatum are not adequately stimulated those parts get either under- or over-stimulated and lead to Parkinson. However, our latter finding (i.e. \textit{Superior Parietal} part) is a novel finding which asserts that not only the \textit{Basal Ganglia} but also \textit{Superior Parietal} part of the brain play role in Parkinson disease.
\section{Conclusion}
In this work, we successfully could build a machine learning model to diagnose Parkinson in patients using MR-Images. We achieved 100 \% accuracy on the validation and test sets, built a brain heat-map for Parkinson diagnosis and verified that \textit{Basal Ganglia} and \textit{Substantia Nigra} part of brain as already were known by medical experts are important in diagnosis of Parkinson, and for the first time we found out that \textit{Superior Parietal} part on right hemisphere of the brain is also very critical in diagnosis of Parkinson.\\
%

{\footnotesize
\bibliographystyle{ieeetr}
\bibliography{egbib}}

\begin{thebibliography}{10}

\bibitem{pd1}
``Parkinson's disease information page.''
  \url{https://www.ninds.nih.gov/Disorders/All-Disorders/Parkinsons-Disease-Information-Page}.

\bibitem{pd2}
G.~. Disease, I.~Incidence, and P.~Collaborators, ``Global, regional, and
  national incidence, prevalence, and years lived with disability for 310
  diseases and injuries, 1990–2015: a systematic analysis for the global
  burden of disease study 2015.''
  \url{https://www.ncbi.nlm.nih.gov/pmc/articles/PMC5055577/}, 2015.

\bibitem{pd3}
``Parkinson's disease information page.''
  \url{https://www.ninds.nih.gov/Disorders/All-Disorders/Parkinsons-Disease-Information-Page}.

\bibitem{Salv}
C.~Salvatorec, ``Machine learning on brain mri data for differential diagnosis
  of parkinson's disease and progressive supranuclear palsy,''

\bibitem{soheil}
S.~Esmaeilzadeh, O.~Khebzegga, and M.~Moradshahi, ``Clinical parameters
  prediction for gait disorder recognition,'' 2018.

\bibitem{zhang}
L.~zhang, ``Classification of parkinson's disease and essential tremor based on
  structural mri,''

\bibitem{Liu}
M.~Liu, J.~Zhang, E.~Adeli, and D.~Shen, ``Deep multi-task multi-channel
  learning for joint classification and regression of brain status. lecture
  notes in computer science (including subseries lecture notes in artificial
  intelligence and lecture notes in bioinformatics),''

\bibitem{Peng}
B.~Peng, ``A multilevel-roi-features-based machine learning method for
  detection of morphometric biomarkers in parkinson’s disease,''

\bibitem{Ahmed}
M.~N. Ahmed and A.~A. Farag, ``Two-stage neural network for volume segmentation
  of medical images,'' {\em Neural Networks, International Conference on pp
  1373- 1378 vol.3.}

\bibitem{Gil}
D.~Gil and M.~Johnsson, ``Diagnosing parkinson by using artificial neural
  networks and support vector machines,'' {\em Global Journal of Computer
  Science and technology}, vol.~9, pp.~63--71, 2009.

\bibitem{Marek}
K.~Marek, ``The parkinson progression marker initiative (ppmi).,'' {\em
  Progress in Neurobiology}, 2011.

\bibitem{Boesen2004}
K.~Boesen, K.~Rehm, and K.~Schaper, ``{Quantitative comparison of four brain
  extraction algorithms},'' {\em NeuroImage}, vol.~22, no.~3, pp.~1255--1261,
  2004.

\bibitem{Smith2002}
S.~Smith, ``{Fast robust automated brain extraction},'' {\em Human Brain
  Mapping}, 2002.

\bibitem{adam}
P.~iederik and J.~Kingma, ``Adam: A method for stochastic optimization,'' 2014.

\bibitem{James}
J.~Bergstra and Y.~Bengio, ``Random search for hyper-parameter optimization,''
  {\em Journal of Machine Learning Research 13 (2012) 281-305}.

\bibitem{gnorm}
Y.~Wu and K.~He, ``Group normalization.''
  \url{https://github.com/taki0112/Group_Normalization-Tensorflow}, journal:
  arXiv:1803.08494v2 [cs.CV] 24 Apr 2018.

\bibitem{Matthew2013}
R.~F. Matthew D.~Zeiler, ``Visualizing and understanding convolutional
  networks,'' {\em arXiv:1311.2901v3 [cs.CV] 28 Nov 2013}, 2013.

\end{thebibliography}

\end{document}